\def\year{2020}\relax
\documentclass[letterpaper]{article} 
\usepackage{aaai20}  
\usepackage{times}  
\usepackage{helvet} 
\usepackage{courier}  
\usepackage[hyphens]{url}  
\usepackage{graphicx} 
\usepackage{graphicx}  
\usepackage{color}
\usepackage{amssymb}
\usepackage{booktabs}
\usepackage{amsfonts}
\usepackage{amsmath}
\usepackage{algorithm} 
\usepackage{algpseudocode} 

\title{DIANet: Dense-and-Implicit Attention Network}
\author{Zhongzhan Huang\textsuperscript{\rm 1}\thanks{Equal contribution},
Senwei Liang\textsuperscript{\rm 2$*$}, Mingfu Liang\textsuperscript{\rm 3},
Haizhao Yang\textsuperscript{\rm 2,4}\\
\textsuperscript{\rm 1}{New Oriental AI Research Academy}\\
\textsuperscript{\rm 2}{National University of Singapore} \\
\textsuperscript{\rm 3}{Northwestern University}\\
\textsuperscript{\rm 4}{Purdue University}\\
hzz\_dedekinds@foxmail.com,\quad
liangsenwei@u.nus.edu,\\
mingfuliang2020@u.northwestern.edu,\quad
matyh@nus.edu.sg}
\begin{document}
\maketitle

\begin{abstract}
Attention networks have successfully boosted the performance in various vision problems.  Previous works lay emphasis on designing a new attention module and individually plug them into the networks. Our paper proposes a novel-and-simple framework that shares an attention module throughout different network layers to encourage the integration of layer-wise information and this parameter-sharing module is referred as Dense-and-Implicit-Attention (DIA) unit. Many choices of modules can be used in the DIA unit. Since Long Short Term Memory~(LSTM) has a capacity of capturing long-distance dependency, we focus on the case when the DIA unit is the modified LSTM~(refer as DIA-LSTM). Experiments on benchmark datasets show that the DIA-LSTM unit is capable of emphasizing layer-wise feature interrelation and leads to significant improvement of image classification accuracy. We further empirically show that the DIA-LSTM has a strong regularization ability on stabilizing the training of deep networks by the experiments with the removal of skip connections or Batch Normalization~\cite{Ioffe:2015:BNA:3045118.3045167} in the whole residual network. 
\end{abstract}

\section{Introduction}
	Attention, a cognitive process that selectively focuses on a small part of information while neglects other perceivable information~\cite{anderson2005cognitive}, has been used to effectively ease neural networks from learning large information contexts from sentences~\cite{vaswani2017attention,britz2017massive,cheng2016long}, images~\cite{Xu:2015:SAT:3045118.3045336,luong2015effective} and videos~\cite{miech2017learnable}. Especially in computer vision, deep neural networks~(DNNs) incorporated with special operators that mimic the attention mechanism can process informative regions in an image efficiently. These operators are modularized and plugged into networks as attention modules~\cite{hu2018squeeze,woo2018cbam,park2018bam,wang2018non,hu2018gather,cao2019GCNet}.
	\begin{figure}[htbp]
		\centering
		\includegraphics[height=0.9in, width=1.5in]{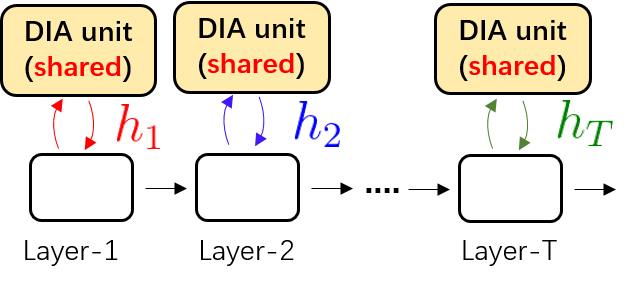}
		\hspace{.1in}
		\includegraphics[height=0.9in, width=1.5in]{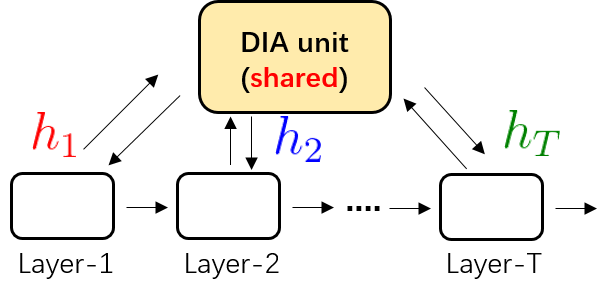}
		\caption{Left: explicit structure of DIANet. Right: implicit connection of DIA unit.}
		\label{imp}
	\end{figure}
	
	Previous works lay emphasis on designing a new attention module and individually plug them into networks. Generally, the attention module can be divided into three parts: extraction, processing and recalibration. First, the added plug-in module extracts internal features of a network which can be squeezed channel-wise information~\cite{hu2018squeeze,li2019selective} or spatial information~\cite{wang2018non,woo2018cbam,park2018bam}. Next, the module processes the extraction and generates a mask to measure the importance of the features via fully connected layer~\cite{hu2018squeeze}, convolution layer~\cite{wang2018non}. Last, the mask is applied to recalibrate the features. Previous works focus on designing effective ways to process the extracted features. There is one obvious common ground where the attention modules are individually plugged into each layer throughout DNNs~\cite{hu2018squeeze,woo2018cbam,park2018bam,wang2018non}.
	
\begin{figure*}[h]
		\centering
		\includegraphics[height=1in, width=5in]{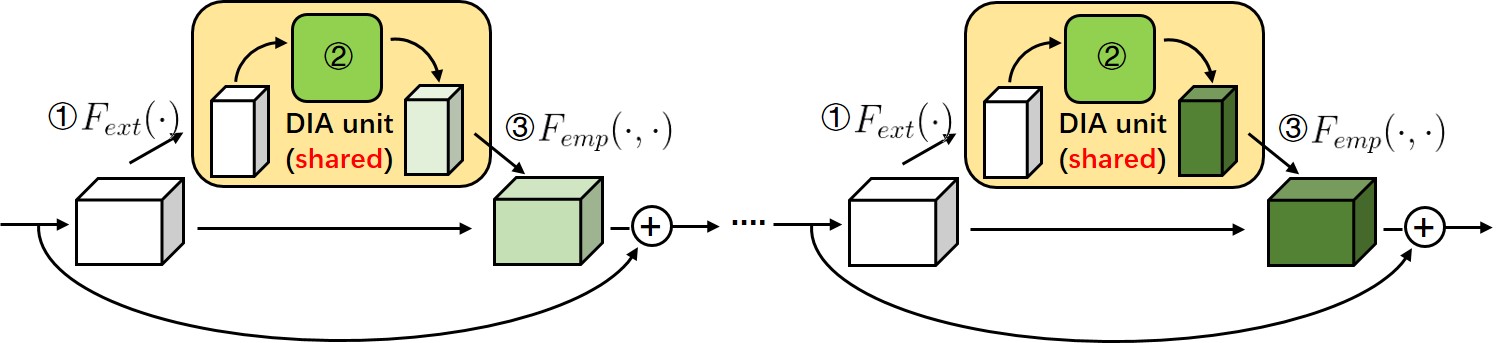}
		\caption{DIA units. $F_{ext}$ means the operation for extracting different scales of features. $F_{emp}$ means the operation for emphasizing features.}
		\label{DIA unit framework}
	\end{figure*}

\textbf{Our Framework.} Differently, we proposes a novel-and-simple framework that shares an attention module throughout different network layers to encourage the integration of layer-wise information and this parameter-sharing module is referred as Dense-and-Implicit-Attention (DIA) unit. The structure and computation flow of a DIA unit is visualized in Figure~\ref{DIA unit framework}. There are also three parts: extraction~({\small{\textcircled{\tiny{1}}}}), processing~({\small{\textcircled{\tiny{2}}}}) and recalibration~({\small{\textcircled{\tiny{3}}}}) in the DIA unit. The {\small{\textcircled{\tiny{2}}}} is the main module in the DIA unit to model network attention and is the key innovation of the proposed method where the parameters of the attention module is shared. 
\textbf{Characteristics and Advantages.} (1) As shown in Figure~\ref{DIA unit framework}, the DIA unit is placed parallel to the network backbone, and it is shared with all the layers in the same stage~(the collection of successive layers with same spatial size, as defined in~\cite{he2016deep}) to improve the interaction of layers at different depth. (2) As the DIA unit is shared, the number of parameter increment from the DIA unit remains roughly constant as the depth of the network increases.

We show the feasibility of our framework by applying SE module~\cite{hu2018squeeze} in DIA unit. SE module, a representative of attention mechanism, is used for each block individually in its original design. In our framework, we share the same SE module~(refer as DIA-SE) throughout all layers in the same stage. It is easy to see that DIA-SE has the same computation cost as SE, but in Table~\ref{tab:share}, DIA-SE has better generalization and smaller parameter increment.

\begin{table}[htbp]
  \centering
    \begin{tabular}{|c|c|c|}
    \toprule
       model   & \#P~(M) & top1-acc. \\
    \midrule
    Org   & 1.73  & 73.43$_{(\pm0.43)}$ \\
    SE    & 1.93  & 75.03$_{(\pm0.33)}$ \\
    DIA-SE & \textbf{1.74}  & \textbf{75.74}$_{(\pm0.41)}$ \\
    \bottomrule
    \end{tabular}%
    \caption{Testing accuracy~(mean $\pm$ std\%) on CIFAR100 and ResNet164 with different attention modules. ''Org'' means the original backbone of ResNet164. \#P~(M) means the number of parameters~(million).}
      \label{tab:share}%
\end{table}%

	 \textbf{Implicit and Dense Connection.} We illustrate how the DIA unit connects all layers in the same stage implicitly and densely. Consider a stage consisting many layers in Figure \ref{imp} (Left). It is an explicit structure with a DIA unit and one layer seems not to connect the other layers except the network backbone. In fact, the different layers use the parameter-sharing attention module and the layer-wise information jointly influences the update of learnable parameters in the module, which causes implicit connections between layers with the help of the shared DIA unit as in Figure~\ref{imp} (Right). Since there is communication between every pair of layers, the connections over all layers are dense.
	 
	\subsection{DIA-LSTM} 
	The idea of parameter sharing also used in Recurrent Neural Network (RNN) to capture contextual information so we consider apply RNN in our framework to model the layer-wise interrelation. Since Long Short Term Memory~(LSTM)~\cite{hochreiter1997long} is capable of capturing long-distance dependency, we mainly focus on the case when we use LSTM in DIA unit~(DIA-LSTM) and the remainder of our paper studies DIA-LSTM.
	
\begin{figure*}[htbp]
		\centering
		\includegraphics[height=1.3in, width=4.4in]{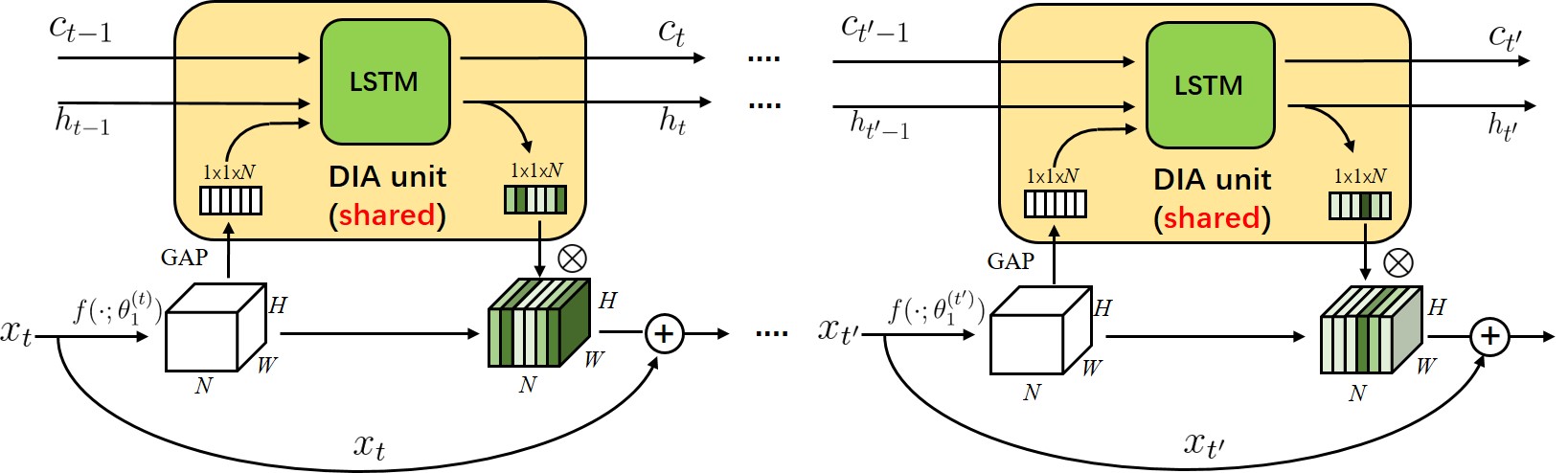}
		\caption{The showcase of DIA-LSTM. In the LSTM cell, $c_t$ is the cell state vector and $h_t$ is the hidden state vector. GAP means global average pool over channels and $\otimes$ means channel-wise multiplication.}
		\label{DIANet structure}
	\end{figure*}

    Figure~\ref{DIANet structure} is the showcase of DIA-LSTM. A global average pooling~(GAP) layer (as the {\small{\textcircled{\tiny{1}}}} in Figure~\ref{DIA unit framework}) is used to extract global information from current layer. A LSTM module (as the {\small{\textcircled{\tiny{2}}}} in Figure~\ref{DIA unit framework}) is used to integrate multi-scale information and there are three inputs passed to the LSTM: the extracted global information from current raw feature map, the hidden state vector $h_{t-1}$, and cell state vector $c_{t-1}$ from previous layers. Then the LSTM outputs the new hidden state vector $h_{t}$ and the new cell state vector $c_{t}$. The cell state vector $c_{t}$ stores the information from the $t^{th}$ layer and its preceding layers. The new hidden state vector $h_{t}$~(dubbed as attention vector in our work) is then applied back to the raw feature map by channel-wise multiplication (as the {\small{\textcircled{\tiny{3}}}} in Figure \ref{DIA unit framework}) to recalibrate the feature. 
	
	The LSTM in the DIA unit plays a role to bridge the current layer and preceding layers such that the DIA unit can adaptively learn the non-linearity relationship between features in two different dimensions. The first dimension of features is the internal information of the current layer. The second dimension represents the outer information, regarded as layer-wise information, from the preceding layers. The non-linearity relationship between these two dimensions will benefit attention modeling for the current layer. The multiple dimension modeling enables DIA-LSTM to have regularization effect.  
	\subsection{Our contribution}
	We summary our contribution as followed,
	\begin{enumerate}
	    \item We proposes a novel-and-simple framework that shares an attention module throughout different network layers to encourage the integration of layer-wise information.
	    \item We propose incorporating LSTM in DIA unit~(DIA-LSTM) and show the effectiveness of DIA-LSTM for image classification by conducting experiments on benchmark datasets and popular networks.
	\end{enumerate}
	
	\section{Related Works} 
	\label{relatedwork}
	\textbf{Attention Mechanism in Computer Vision.} \cite{mnih2014recurrent,zhao2017diversified} use attention mechanism in image classification via utilizing a recurrent neural network to select and process local regions at high resolution sequentially. Concurrent attention-based methods tend to construct operation modules to capture non-local information in an image~\cite{wang2018non,cao2019GCNet}, and model the interrelationship between channel-wise features~\cite{hu2018squeeze,hu2018gather}. The combination of multi-level attentions are also widely studied~\cite{park2018bam,woo2018cbam,DBLP:journals/corr/abs-1904-04402,Wang_2017_CVPR}. Prior works~\cite{wang2018non,cao2019GCNet,hu2018squeeze,hu2018gather,park2018bam,woo2018cbam,DBLP:journals/corr/abs-1904-04402} usually insert an attention module in each layer individually. In this work, the DIA unit is innovatively shared for all the layers in the same stage of the network, and the existing attention modules can be composited into the DIA unit readily. Besides, we adopt a global average pooling in part {\small{\textcircled{\tiny{1}}}} to extract global information and a channel-wise multiplication in part {\small{\textcircled{\tiny{3}}}} to recalibrate features, which is similar to SENet~\cite{hu2018squeeze}. 

	\noindent\textbf{Dense Network Topology.} DenseNet proposed in  \cite{huang2017densely} connects all pairs of layers directly with an identity map. Through reusing features, DenseNet has the advantage of higher parameter efficiency, the better capacity of generalization, and more accessible training than alternative architectures~\cite{lin2013network,he2016deep,srivastava2015highway}. Instead of explicitly dense connections, the DIA unit implicitly links layers at different depth via a shared module and leads to dense connection.
	
	\noindent\textbf{Multi-Dimension Feature Integration.} \cite{wolf2006critical} experimentally analyzes that even the simple aggregation of low-level visual features sampled from wide inception field can be efficient and robust for context representation, which inspires \cite{hu2018squeeze,hu2018gather} to incorporate multi-level features to improve the network representation. \cite{li2016multi} also demonstrates that by biasing the feature response in each convolutional layers using different activation functions, the deeper layer could achieve the better capacity of capturing the abstract pattern in DNN. In DIA unit, the high non-linearity relationship between multi-dimension features are learned and integrated via the LSTM module. 
	\section{Dense-and-Implicit Attention Network}
	\label{DIA}
	In this section, we will formally introduce the DIA-LSTM unit. We use the modified LSTM module in the DIA unit. Afterwards, a DIANet is referred to a network built with DIA-LSTM units. 
	\subsection{Formulation of DIA-LSTM unit}\label{subsec:Formulation of DIANet}
    As shown in Figure~\ref{DIANet structure} when a DIA-LSTM unit is built with a residual network~\cite{he2016deep}, the input of the $t^{th}$ layer is $x_t\in \mathbb{R}^{W\times H \times N}$, where $W, H$ and $N$ mean width, height and the number of channels, respectively. $f(\cdot; \theta_1^{(t)})$ is the residual mapping  at the $t^{th}$ layer with parameters $\theta_1^{(t)}$ as introduced in~\cite{he2016deep}. Let $a_t = f(x_t; \theta_1^{(t)})\in\mathbb{R}^{W\times H \times N}.$ Next, a global average pooling denoted as $\text{GAP}(\cdot)$ is applied to $a_t$ to extract global information from features in the current layer. Then $\text{GAP}(a_t)\in \mathbb{R}^{ N}$ is passed to LSTM along with a hidden state vector $h_{t-1}$ and a cell state vector $c_{t-1}$ ( $h_0$ and $c_0$ are initialized as zero vectors). The LSTM finally generates a current hidden state vector $h_{t}\in \mathbb{R}^{N}$ and a cell state vector $c_{t}\in \mathbb{R}^{N}$ as
	\begin{align}
	(h_t,c_t) = \text{LSTM}(\text{GAP}(a_t), h_{t-1},c_{t-1};\theta_{2}).
	\label{eqn:lstm-dia}
	\end{align}
	 In our model, the hidden state vector $h_t$ is regarded as attention vector to adaptively recalibrate feature maps. We apply channel-wise multiplication $\otimes$ to enhance the importance of features, i.e., $a_t \otimes h_t$ and obtain $x_{t+1}$ after skip connection, i.e., $x_{t+1} = x_t + a_t \otimes h_t$. Table~\ref{DIANet compare SENet} shows the formulation of ResNet, SENet, and DIANet, and Part (b) is the main difference between them. The LSTM module is used repeatedly and shared with different layers in parallel to the network backbone. Therefore the number of parameters $\theta_{2}$ in a LSTM does not depend on the number of layers in the backbone, e.g., $t$. SENet utilizes a attention-module consisted of fully connected layers to model the channel-wise dependency for each layer individually~\cite{hu2018squeeze}. The total number of parameters brought by the add-in modules depends on the number of layers in the backbone and increases with the number of layers. 
	\begin{table*}[htbp]
		\small
		\centering
		\begin{tabular}{|c|c|c|c|}
			\toprule
			&    ResNet   &  SENet     & DIANet (ours) \\
			\midrule
			(a)  &  $a_t = f(x_t;\theta_{1}^{(t)})$ & $a_t = f(x_t;\theta_{1}^{(t)})$ & $a_t = f(x_t;\theta_{1}^{(t)})$  \\
			(b)  &    -   &   $h_t = \text{FC}(\text{GAP}(a_t);\theta_{2}^{(t)})$    &  $(h_t,c_t) = \text{LSTM}(\text{GAP}(a_t), h_{t-1},c_{t-1};\theta_{2})$\\
			(c) & $x_{t+1} = x_t + a_t$      &    $x_{t+1} = x_t + a_t \otimes h_t$   & $x_{t+1} = x_t + a_t \otimes h_t$ \\
			\bottomrule
		\end{tabular}%
		\caption{Formulation for the structure of ResNet, SENet, and DIANet. $f$ is the convolution layer. $\text{FC}$ means fully connected layer and $\text{GAP}$ indicates global average pooling. }
		\label{DIANet compare SENet}%
	\end{table*}%

	\subsection{Modified LSTM Module}
	\label{sec:lstm}
\begin{figure}[h]
		\centering
		\includegraphics[height=2in, width=2.in]{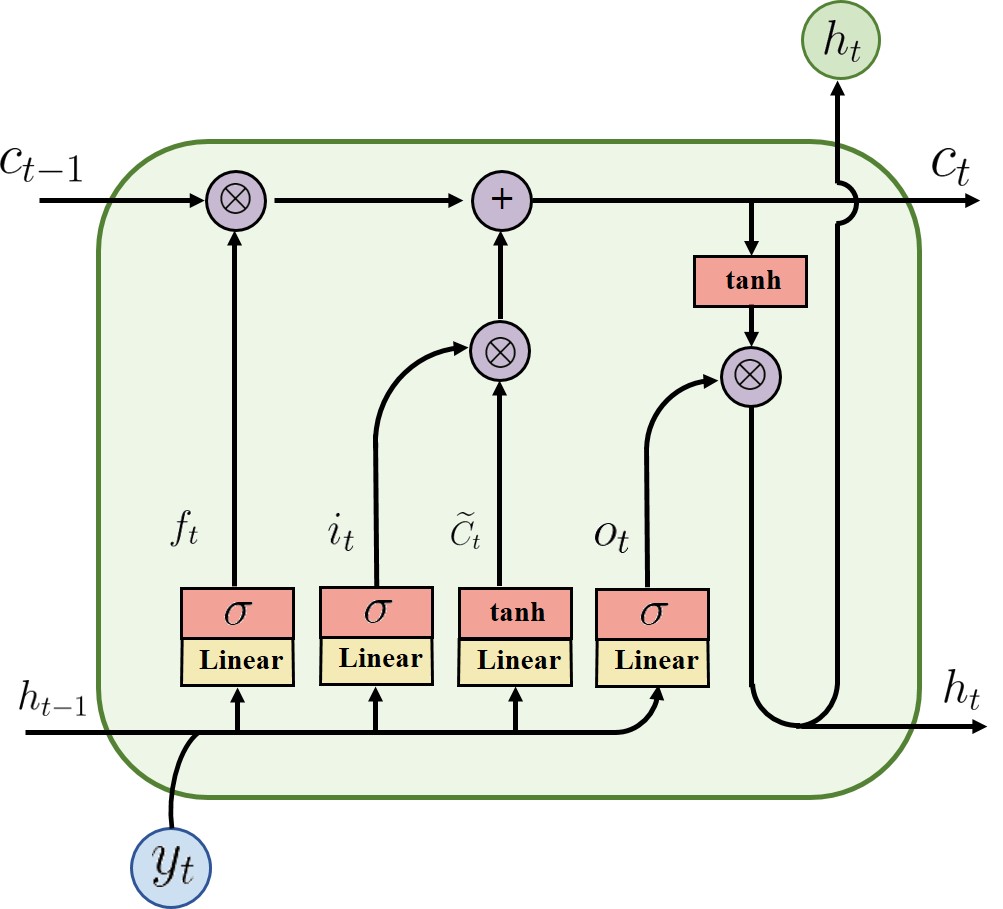}
		\hspace{.5in}
		\includegraphics[height=2in, width=2.in]{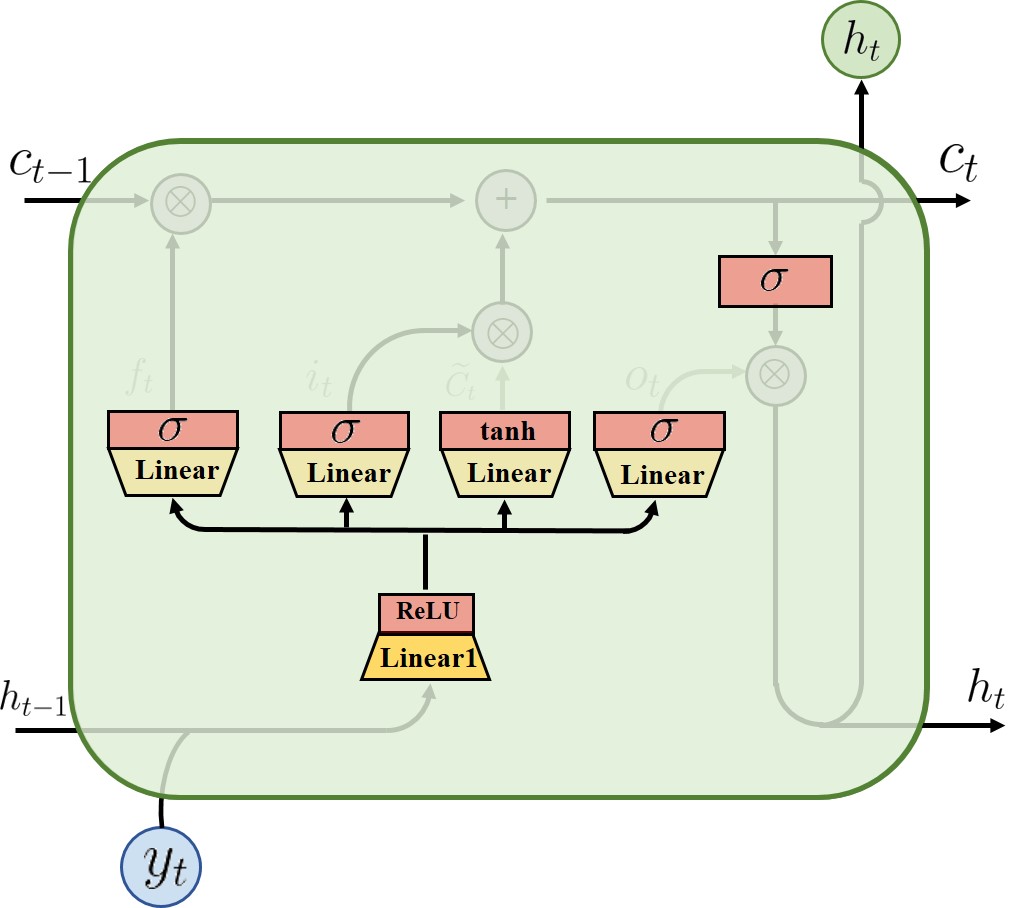}
		\caption{\textbf{Top}. The standard LSTM cell. \textbf{Bottom}. The modified LSTM cell in DIA-LSTM unit. We highlight the modified component in the modified LSTM. ``$\sigma$'' means the sigmoid activation. ``Linear'' means the linear transformation.
		}
		\label{LSTM compare}
	\end{figure}

	Now we introduce the modified LSTM module used in Figure~\ref{DIANet structure}. The design of attention module usually requires the value of the attention vector in range $[0,1]$ and also requires small parameter increment. We conducts some modifications in LSTM module used in DIA-LSTM. As shown in Figure~\ref{LSTM compare}, compared to the standard LSTM~\cite{hochreiter1997long} module, there are two modifications in our purposed LSTM: 1) a shared linear transformation to reduce input dimension of LSTM; 2) a careful selected activation function for better performance. 
	
	\noindent\textbf{(1) Parameter Reduction.} A standard LSTM consists of four linear transformation layers as shown in Figure~\ref{LSTM compare} (Left). Since $y_{t}$, $h_{t-1}$ and $h_{t}$ are of the same dimension $N$, the standard LSTM may cause $8N^2$ parameter increment as shown in Appendix. When the number of channels is large, e.g., $N=2^{10}$, the parameter increment of added-in LSTM module in the DIA unit will be over 8 million, which can hardly be tolerated. 
	
	As shown in Figure \ref{LSTM compare}~(Top), $h_{t-1}$ and $y_{t}$ are passed to four linear transformation layers with the same input and output dimension $N$. In the DIA-LSTM, a linear transformation layer (denoted as ``Linear1'' in Figure \ref{LSTM compare} (Bottom)) with a smaller output dimension are applied to $h_{t-1}$ and $y_{t}$. We use reduction ratio $r$ in the Linear1. Specifically, we reduce the dimension of the input  from $1 \times 1 \times N$ to $1 \times 1 \times N/r$ and then apply the ReLU activation function to increase non-linearity in this module. The dimension of the output from ReLU function are changed back to $1 \times 1 \times N$ when the output is passed to those four linear transformation functions. This modification can enhance the relationship between the inputs for different parts in DIA-LSTM and also effectively reduce the number of parameters by sharing a linear transformation for dimension reduction. The number of parameter increment reduces from $8N^2 $ to $10 N^2/r$ as shown in the Appendix, and we find that when we choose an appropriate reduction ratio $r$, we can make a better trade-off between parameter reduction and the performance of DIANet. Further experimental results will be discussed in the ablation study later.

	\noindent\textbf{(2) Activation Function.} Sigmoid function ($\sigma(z) = 1/(1+e^{-z})$) is used in many attention-based methods like SENet~\cite{hu2018squeeze}, CBAM~\cite{woo2018cbam} to generate attention maps as a gate mechanism. As shown in Figure~\ref{LSTM compare}~(Bottom), we change the activation function of the output layer from Tanh to Sigmoid. Further discussion will be presented in ablation study.
	\section{Experiments}
	\label{experiments}
	In this section, we evaluate the performance of the DIA-LSTM unit in image classification task and empirically demonstrate its effectiveness. We conduct experiments on popular networks for benchmark datasets. Since SENet~\cite{hu2018squeeze} is also a channel-specific attention model, we compare DIANet with SENet. For a fair comparison, we adjust the reduction ratio such that the number of parameters of DIANet is similar to that of SENet. 
	
	\textbf{Dataset and Model.} We conduct experiments on CIFAR10, CIFAR100~\cite{cifar}, and ImageNet 2012~\cite{ILSVRC15} using ResNet~\cite{he2016deep}, PreResNet~\cite{he2016identity}, WRN~\cite{wrn} and ResNeXt~\cite{xie2017aggregated}. CIAFR10 or CIFAR100 has 50k train images and 10k test images of size 32 by 32, but has 10 and 100 classes respectively. ImageNet 2012~\cite{ILSVRC15} comprises 1.28 million training and 50k validation images from 1000 classes, and the random cropping of size 224 by 224 is used in our experiments. The details can be found in Appendix. 
	
	\textbf{Image Classification.} As shown in Table \ref{classification result}, DIANet improves the testing accuracy significantly over the original networks and consistently comparing with SENet for different datasets. In particular, the performance improvement of the ResNet with the DIA unit is most remarkable. Due to the popularity of ResNet, the DIA unit may be applied in other computer vision tasks.	
		\begin{table*}[h]
		\centering
		\small
		\begin{tabular}{|l|c|c|c|c|c|c|c|c|}
			\toprule
			& Dataset & \multicolumn{2}{c|}{original} & \multicolumn{2}{c|}{SENet} & \multicolumn{3}{c|}{DIANet} \\
			\cmidrule{3-9}          &       & $\#$P(M) & top1-acc. & $\#$P(M) & top1-acc. & $\#$P(M) & top1-acc. & $r$ \\
			\midrule
			ResNet164 & CIFAR100 & 1.73  & 73.43  & 1.93  & 75.03  & 1.95  & \textbf{76.67 } & 4 \\
			PreResNet164 & CIFAR100 & 1.73  & 76.53  & 1.92  & 77.41  & 1.96  & \textbf{78.20 } & 4 \\
			WRN52-4 & CIFAR100 & 12.07  & 79.75  & 12.42  & 80.35  & 12.30  & \textbf{80.99 } & 4 \\
			ResNext101,8x32 & CIFAR100 & 32.14  & 81.18  & 34.03  & 82.45  & 33.01  & \textbf{82.46 } & 4 \\
			\midrule
			ResNet164 & CIFAR10 & 1.70  & 93.54  & 1.91  & 94.27  & 1.92  & \textbf{94.58 } & 4 \\
			PreResNet164 & CIFAR10 & 1.70  & 95.01  & 1.90  & 95.18  & 1.94  & \textbf{95.23 } & 4 \\
			WRN52-4 & CIFAR10 & 12.05  & 95.96  & 12.40  & 95.95  & 12.28  & \textbf{96.17 } & 4 \\
			ResNext101,8x32 & CIFAR10 & 32.09  & 95.73  & 33.98  & 96.09  & 32.96  & \textbf{96.24 } & 4 \\
			\midrule
			ResNet34 & ImageNet & 21.81     & 73.93  & 21.97    & 74.39  & 21.98     & \textbf{74.60 } & 20 \\
			ResNet50 & ImageNet & 25.58     & 76.01  & 28.09    & 76.61  &  28.38   & \textbf{77.24 } & 20 \\
			ResNet152 & ImageNet & 60.27     & 77.58  & 66.82     & 78.36  &  65.85    & \textbf{78.87 } & 10 \\
			ResNext50,32x4 & ImageNet & 25.03     & 77.19  &  27.56    & 78.04  &  27.83    & \textbf{78.32 } & 20 \\
			\bottomrule
		\end{tabular}%
		\caption{Testing accuracy (\%) on CIFAR10, CIFAR100 and ImageNet 2012. ``$\#$P(M)'' means the number of parameters (million). The rightmost ``$r$'' indicates the reduction ratio of DIANet.}
		\label{classification result}%
	\end{table*}%

	\section{Ablation Study}
	\label{ablationstudy}
	In this section, we conduct ablation experiments to explore how to better embed DIA-LSTM units in different neural network structures and gain a deeper understanding of the role of components in the unit. All experiments are performed on CIFAR100 with ResNet. For simplicity, DIANet164 is denoted as a 164-layer ResNet built with DIA-LSTM units.	
	
	\noindent\textbf{Reduction ratio.} The reduction ratio is the only hyperparameter in DIANet. The main advantage of our model is to improve the generalization ability with a light parameter increment. The smaller reduction ratio causes a higher parameter increment and model complexity. This part investigates the trade-off between the model complexity and performance. As shown in Table~\ref{tab:reduction ratio}, the number of parameters of the DIANets decreases with the increasing reduction ratio, but the testing accuracy declines slightly, which suggests that the model performance is not sensitive to the reduction ratio. In the case of $r=16$, the DIANet164 has 0.05M parameter increment compared to the original ResNet164 but the testing accuracy of the DIANet164 is 76.50\% while that of the ResNet164 is 73.43\%. 
	\begin{table}[htbp]
	    \centering
	\small
			\centering
			\begin{tabular}{|c|c|c|}
				\midrule
				Ratio $r$ & $\#$P(M) & top1-acc. \\
				\midrule
				1  & 2.59$_{(+0.86)}$ & 76.88  \\
				4  & 1.95$_{(+0.22)}$  & 76.67  \\
				8  & 1.84$_{(+0.11)}$  & 76.42  \\
				16  & 1.78$_{(+0.05)}$  & 76.50  \\
				\bottomrule
			\end{tabular}%
			\caption{Test accuracy (\%) with different reduction ratio on CIFAR100 with ResNet164.}
			\label{tab:reduction ratio}%
	\end{table}{}
	\begin{figure*}[h]

		\centering
		\includegraphics[height=1.4in, width=5in]{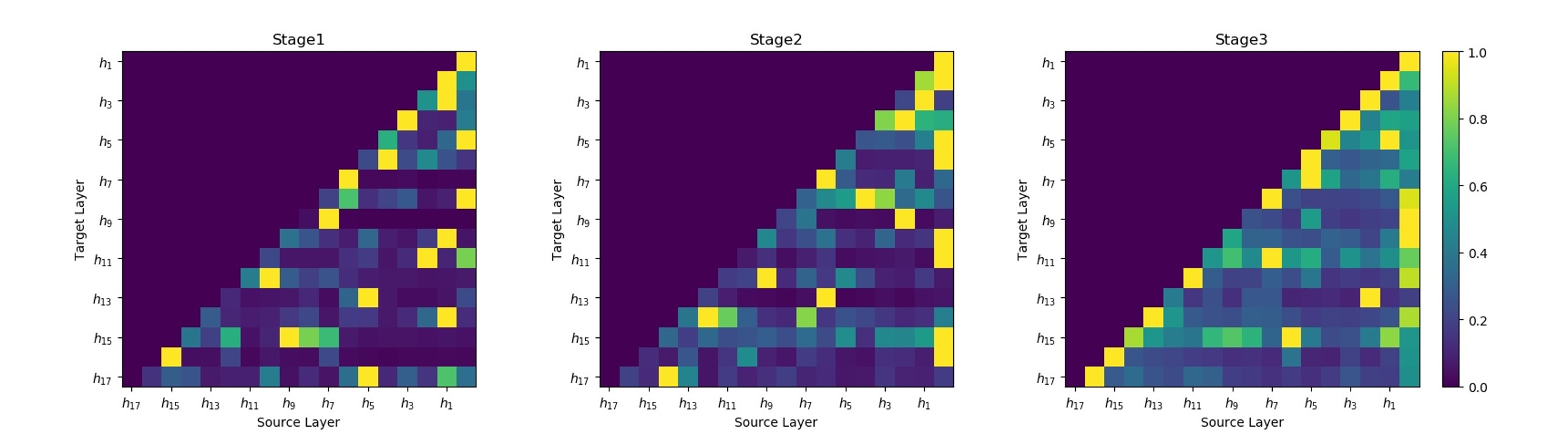}
		\caption{Visualization of feature integration for each stage by random forest. Each row presents the importance of source layers $h_n,1\leq n < t$ contributing to the target layer $h_t$.}
		\label{features integration for each stage}
	\end{figure*}
	
	\noindent\textbf{The depth of the neural network.} Generally, in practice, DNNs with a larger number of parameters do not guarantee sufficient performance improvement. Deeper networks may contain extreme feature and parameter redundancy~\cite{huang2017densely}. Therefore,  designing a new structure of deep neural networks~\cite{he2016deep,huang2017densely,srivastava2015training,hu2018squeeze,hu2018gather,wang2018non} is necessary. Since DIA units change the topology of DNN backbones, evaluating the effectiveness of DIANet structure is of great importance. Here we investigate how the depth of DNNs influences DIANets in two aspects: (1) the performance of DIANets compared to SENets of various depth; (2) the parameter increment of DIANets. 
	
	The results in Table \ref{tab:DIANet with different depth1} show that as the depth increases from 83 to 407 layers, the DIANet with a smaller number of parameters can achieve higher classification accuracy than the SENet. Moreover, the DIANet83 can achieve a similar performance as the SENet245, and DIANet164 outperforms all the SENets with at least 1.13\% and at most 58.8\% parameter reduction. They imply that the DIANet is of higher parameter efficiency than SENet. The results also suggest that: for DIANet, as shown in Figure~\ref{DIANet structure}, the DIA-LSTM unit will pass more layers recurrently with a deeper depth. The DIA-LSTM can handle the interrelationship between the information of different layers in much deeper DNN and figure out the long-distance dependency between layers. 
		\begin{table}[h]
	  		\small
			\centering
			\begin{tabular}{|l|c|c|c|c|}
				\toprule
				\multicolumn{1}{|c|}{CIFAR-100} & \multicolumn{2}{c|}{SENet} & \multicolumn{2}{c|}{DIANet($r=4$)} \\
				\cmidrule{2-5}    \multicolumn{1}{|c|}{Depth} & $\#$P(M) & top1-acc.  & $\#$P(M) & top1-acc. \\
				\midrule
				ResNet83 & 0.99  & 74.67  & 1.11$_{(+0.12)}$ & 75.02  \\
				ResNet164 & 1.93  & 75.03  & 1.95$_{(+0.02)}$ & 76.67  \\
				ResNet245 & 2.87  & 75.03  & 2.78$_{(-0.09)}$ & 76.79  \\
				ResNet407 & 4.74  & 75.54  & 4.45$_{(-0.29)}$ & 76.98  \\
				\bottomrule
			\end{tabular}%
			\caption{Test accuracy (\%) with ResNet of different depth on CIFAR100.}
			\label{tab:DIANet with different depth1}%
	\end{table}{}	

	\noindent\textbf{Activation function and the number of stacking LSTM cells.} We choose different activation functions in the output layer of LSTM in Figure~\ref{LSTM compare}~(Bottom) and different numbers of stacking LSTM cells to explore the effects of these two factors. 
	In Table \ref{tab:activation and number of unit}, we find that the performance has been significantly improved after replacing tanh in the standard LSTM with sigmoid. As shown in Figure~\ref{LSTM compare} (Bottom), this activation function is located in the output layer and directly changes the effect of memory unit $c_t$ on the output of the output gate. In fact, the sigmoid function is used in many attention-based methods like SENet as a gate mechanism. The test accuracy of different choices of LSTM activation functions in Table~\ref{tab:activation and number of unit} shows that sigmoid better helps LSTM as a gate to rescale channel features. Table~12 in the SENet paper~\cite{hu2018squeeze} shows the performance of different activation functions like: sigmoid $>$ tanh $>$ ReLU~(bigger is better), which coincides to our reported results.
	
	When we use sigmoid in the output layer of LSTM, the increasing number of stacking LSTM cells does not necessarily lead to performance improvement but may lead to performance degradation. However, when we choose tanh, the situation is different. 
	It suggest that, through the stacking of LSTM cells, the scale of the information flow among them is changed, which may effect the performance.
	
	\begin{table}[htbp]
		\small
		\centering
		\begin{tabular}{|c|c|c|c|}
			\toprule
			$\#$P(M) & Activation & $\#$LSTM cells& top1-acc. \\
			\midrule
			 1.95  & sigmoid & 1  & 76.67  \\
			 1.95  & tanh  & 1  & 75.24  \\
			 1.95  & ReLU & 1  & 74.62  \\
			 3.33  & sigmoid & 3  & 75.20  \\
			 3.33  & tanh  & 3  & 76.47  \\
			\bottomrule
		\end{tabular}
		\caption{Test accuracy (\%) with DIANet164 of different activation function at the output layer in the modified LSTM and different number of stacking LSTM cells on CIFAR100.}
		\label{tab:activation and number of unit}
	\end{table}	
	
	\section{Analysis}
	\label{analysis}
	
	In this section, we study some properties of DIANet, including feature integration and regularization effect on stabilizing training. Firstly, the layers are connected by DIA-LSTM unit in DIANet and we can use the random forest model~\cite{Gregorutti2017Correlation} to visualize how the current layer depends on the preceding layers. Secondly, we study the stabilizing training effect of DIANet by removing all the Batch Normalization~\cite{Ioffe:2015:BNA:3045118.3045167} or the skip connection in the residual networks.
	
	\subsection{Feature Integration}
	Here we try to understand the dense connection from the numerical perspective. As shown in Figure~\ref{DIANet structure} and~\ref{imp}, the DIA-LSTM bridges the connections between layers by propagating the information forward through $h_t$ and $c_t$. Moreover, $h_t$ at different layers are also integrating with $h_{t'},1\leq t'<t$ in DIA-LSTM. Notably, $h_t$ is applied directly to the features in the network at each layer $t$. Therefore the relationship between $h_t$ at different layers somehow reflects connection degree of different layers. We explore the nonlinear relationship between the hidden state $h_t$ of DIA-LSTM and the preceding hidden state $h_{t-1},h_{t-2},...,h_1$, and visualize how the information coming from $h_{t-1},h_{t-2},...,h_1$ contributes to $h_t$. To reveal this relationship, we consider using the random forest to visualize variable importance. The random forest can return the contributions of input variables to the output separately in the form of importance measure, e.g., Gini importance~\cite{Gregorutti2017Correlation}. The computation details of Gini importance can be referred to the Appendix. Take $h_n,1\leq n < t$ as input variables and $h_t$ as output variable, we can get the Gini importance of each variable $h_n,1\leq n < t$. ResNet164 contains three stages, and each stage consists of 18 layers. We conduct three Gini importance computation to each stage separately. As shown in Figure~\ref{features integration for each stage}, each row presents the importance of source layers $h_n,1\leq n < t$ contributing to the target layer $h_t$. In each sub-graph of Figure~\ref{features integration for each stage}, the diversity of variable importance distribution indicates that the current layer uses the information of the preceding layers. The interaction between shallow and deep layers in the same stage reveals the effect of implicitly dense connection. In particular, taking $h_{17}$ in stage 1~(the last row) as an example, $h_{16}$ or $h_{15}$ does not intuitively provide the most information for $h_{17}$, but $h_5$ does. We conclude that the DIA unit can adaptively integrate information between multiple layers.
	\begin{table}[h]
			\small
			\centering
			\begin{tabular}{|c|c|c|c|c|}
				\toprule
				stage removed& $\#$P(M) & $\#$P(M)$\downarrow$ & top1-acc. & top1-acc.$\downarrow$ \\
				\midrule
				stage1 & 1.94  & 0.01  & 76.27 & 0.40 \\
				stage2 & 1.90   & 0.05  & 76.25 & 0.42 \\
				stage3 & 1.78  & 0.17  & 75.40  & 1.27 \\
				\bottomrule
			\end{tabular}%
    \caption{The test accuracy (\%) of DIANet164 with the removal of DIA-LSTM unit in different stage.}
    \label{tab:remove stage}
\end{table}
\begin{table*}[ht]
		\small
		\centering
		\begin{tabular}{|l|c|c|c|c|c|c|}
			\toprule
			& \multicolumn{2}{c|}{original} & \multicolumn{2}{c|}{SENet} & \multicolumn{2}{c|}{DIANet$(r=16)$} \\
			\cmidrule{2-7}          & $\#$P(M) & top1-acc. & $\#$P(M) & top1-acc. & $\#$P(M) & top1-acc. \\
			\midrule
			ResNet83 & 0.88  & nan   & 0.98  & nan   & 0.94  & \textbf{70.58 } \\
			ResNet164 & 1.70  & nan   & 1.91  & nan   & 1.76  & \textbf{72.36 } \\
			ResNet245 & 2.53  & nan   & 2.83  & nan   & 2.58  & \textbf{72.35 } \\
			ResNet326 & 3.35  & nan   & 3.75  & nan   & 3.41  & nan \\
			\bottomrule
		\end{tabular}%
		\caption{Testing accuracy (\%). We train models of different depth without BN on CIFAR-100. ``nan'' indicates the numerical explosion. }
		\label{tab:withoutbn}%
	\end{table*}%
	\begin{figure*}[h]
    \centering
    \includegraphics[height=1.2in, width=4in]{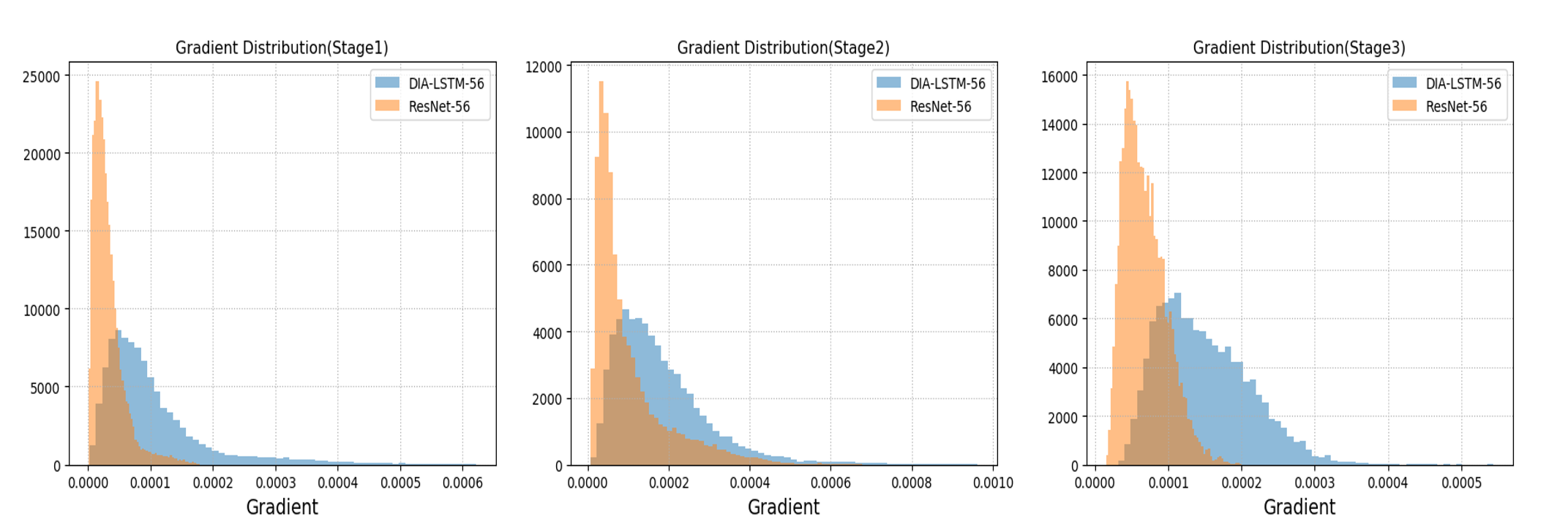}
    \caption{The distribution of gradient in each stage of ResNet56 without all the skip connections.}
    \label{fig:gradient}
\end{figure*}
	Moreover, in Figure~\ref{features integration for each stage}~(stage 3), the information interaction with previous layers in stage 3 is more intense and frequent than that of the first two stages. Correspondingly, as shown in Table~\ref{tab:remove stage}, in the experiments when we remove the DIA-LSTM unit in stage 3, the classification accuracy decreases from 76.67 to 75.40. However, when it in stage 1 or 2 is removed, the performance degradation is very similar, falling to 76.27 and 76.25 respectively. Also note that for DIANet, the number of parameter increment in stage 2 is larger than that of stage 1. It implies that the significant performance degradation after the removal of stage 3 may be not only due to the reduction of the number of parameters but due to the lack of dense feature integration.

	\subsection{The effect on stabilizing training}
	\noindent\textbf{Removal of Batch Normalization.} Small changes in shallower hidden layers may be amplified as the information propagates within the deep architecture and sometimes result in a numerical explosion. Batch Normalization (BN)~\cite{Ioffe:2015:BNA:3045118.3045167} is widely used in the deep networks since it stabilizes the training by normalization the input of each layer. DIA-LSTM unit recalibrates the feature maps by channel-wise multiplication, which plays a role of scaling similar to BN. As shown in Table~\ref{tab:withoutbn}, different models trained with varying depth in CIFAR100 and BNs are removed in these networks. The experiments are conducted on a single GPU with batch size 128 and initial learning rate 0.1. Both the original ResNet, SENet face problem of numerical explosion without BN while the DIANet can be trained with depth up to 245. In Table~\ref{tab:withoutbn}, at the same depth, SENet has larger number of parameters than DIANet but still comes to numerical explosion without BN, which means that the number of parameter is not the case for stabilization of training but sharing mechanism we proposed may be the case. Besides, comparing with Table~\ref{tab:DIANet with different depth1}, the testing accuracy of DIANet without BN still can keep up to 70\%. The scaling learned by DIANet integrates the information from preceding layers and enables the network to choose a better scaling for feature maps of current layer. 
	
	\noindent\textbf{Removal of skip connection.} The skip connection has become a necessary structure for training DNNs~\cite{he2016identity}. Without skip connection, the DNN is hard to train due to the reasons like the gradient vanishing~\cite{bengio1994learning,glorot2010understanding,srivastava2015training}. We conduct the experiment where all the skip connections are removed in ResNet56 and count the absolute value of gradient at the output tensor of each stage. As shown in Figure~\ref{fig:gradient} which presents the gradient distribution with all skip connection removal, DIANet~(blue) obviously enlarges the mean and variance of the gradient distribution, which enables larger absolute value and diversity of gradient and relieves gradient degradation to some extent.

	\noindent\textbf{Without data augment.} Explicit dense connections may help bring more efficient usage of parameters, which makes the neural network less prone to overfit~\cite{huang2017densely}. Although the dense connections in DIA-LSTM are implicit, the DIANet still shows the ability to reduce overfitting. To verify it, We train the models without data augment to reduce the influence of regularization from data augment. As shown in Table~\ref{tab:remove stage and without aug}, DIANet achieves lower testing error than ResNet164 and SENet. To some extent, the implicit and dense structure of DIANet may have regularization effect. 
	\begin{table}
			\small
			\centering
			\begin{tabular}{|c|c|c|}
				\toprule
				Models    & CIFAR-10  & CIFAR-100 \\
				\midrule
				ResNet164 & 87.32 &60.92 \\
				SENet & 88.30     & 62.91 \\
				DIANet & \textbf{89.25}     & \textbf{66.73} \\
				\bottomrule
			\end{tabular}%
		\caption{Test accuracy (\%) of the models without data augment with ResNet164.}
		\label{tab:remove stage and without aug}%
    \end{table}{}	
	\section{Conclusion}
	\label{conculsion}
	In this paper, we proposes a novel-and-simple framework that shares an attention module throughout different network layers to encourage the integration of layer-wise information. The parameter-sharing module is called Dense-and-Implicit Attention (DIA) unit. 
	We propose incorporating LSTM in DIA unit~(DIA-LSTM) and show the effectiveness of DIA-LSTM for image classification by conducting experiments on benchmark datasets and popular networks.
	We further empirically show that the DIA-LSTM has a strong regularization ability on stabilizing the training of deep networks by the experiments with the removal of skip connections or Batch Normalization~\cite{Ioffe:2015:BNA:3045118.3045167} in the whole residual network.
	\small
 	\section{Acknowledgments}
	
 	S. Liang and H. Yang gratefully acknowledge the support of National Supercomputing Center (NSCC) SINGAPORE and High Performance Computing (HPC) of National University of Singapore for providing computational resources, and the support of NVIDIA Corporation with the donation of the Titan Xp GPU used for this research. Sincerely thank Xin Wang from Tsinghua University for providing personal computing resource. H. Yang thanks the support of the start-up grant by the Department of Mathematics at the National University of Singaporet, the Ministry of Education in Singapore for the grant MOE2018-T2-2-147.


	\appendix
	\section{Introdcution of Implementation detail}
	The hyper-parameter settings of  CIFAR and ImageNet are shown in Table~\ref{tab:cifar} and Table~\ref{tab:imagenet} respectively. 
	\begin{table*}[htbp]
		\small
		\centering
		\begin{tabular}{|c|c|c|c|c|}
			\toprule
			& ResNet164 & PreResNet164 & WRN52-4 & ResNext101-8x32 \\
			\midrule
			Batch size & 128   & 128   & 128   & 128 \\
			Epoch & 180   & 164   & 200   & 300 \\
			Optimizer & SGD(0.9) & SGD(0.9) & SGD(0.9) & SGD(0.9) \\
			depth & 164   & 164   & 52    & 101 \\
			schedule & 60/120 & 81/122 & 80/120/160 & 150/225 \\
			wd    & 1.00E-04 & 1.00E-04 & 5.00E-04 & 5.00E-04 \\
			gamma & 0.1   & 0.1   & 0.2   & 0.1 \\
			widen-factor & -     & -     & 4     & 4 \\
			cardinality & -     & -     & -     & 8 \\
			lr    & 0.1   & 0.1   & 0.1   & 0.1 \\
			$F_{ext}(\cdot)$ & GAP   & BN+GAP & BN+GAP & GAP \\
			drop  & -     & -     & 0.3   & - \\
			\bottomrule
		\end{tabular}%
		\caption{Implementation detail for \textbf{CIFAR10/100} image classification. Normalization and standard data augmentation (random cropping and horizontal flipping) are applied to the training data. GAP and BN denote Global Average Pooling and Batch Normalization separately. }
		\label{tab:cifar}%
		\vspace{-0.2cm}
	\end{table*}%

	\begin{table*}[htbp]
		
		\centering
		\begin{tabular}{|c|c|c|c|c|}
			\toprule
			& ResNet34 & ResNet50 & ResNet152 & ResNext50-32x4 \\
			\midrule
			Batch size & 256   & 256   & 256   & 256 \\
			Epoch & 120   & 120   & 120   & 120 \\
			Optimizer & SGD(0.9) & SGD(0.9) & SGD(0.9) & SGD(0.9) \\
			depth & 34    & 50    & 152   & 50 \\
			schedule & 30/60/90 & 30/60/90 & 30/60/90 & 30/60/90 \\
			wd    & 1.00E-04 & 1.00E-04 & 1.00E-04 & 1.00E-04 \\
			gamma & 0.1   & 0.1   & 0.1   & 0.1 \\
			lr    & 0.1   & 0.1   & 0.1   & 0.1 \\
			$F_{ext}(\cdot)$ & GAP   & GAP   & GAP   & GAP \\
			\bottomrule
		\end{tabular}%
		\caption{Implementation detail for \textbf{ImageNet 2012} image classification. Normalization and standard data augmentation (random cropping and horizontal flipping) are applied to the training data. The random cropping of size 224 by 224 is used in these experiments. GAP denote Global Average Pooling .}
		\vspace{-0.18cm}
			\label{tab:imagenet}%
	\end{table*}%
	
	\begin{table*}[htbp]
		\small
		\centering
		\begin{tabular}{|l|l|}
			\toprule
			Batch size & train batchsize \\
			Epoch & number of total epochs to run \\
			Optimizer & Optimizer \\
			depth & the depth of the network \\
			schedule & Decrease learning rate at these epochs \\
			wd    & weight decay \\
			gamma & learning rate is multiplied by gamma on schedule \\
			widen-factor & Widen factor \\
			cardinality & Model cardinality (group) \\
			lr    & initial learning rate \\
			$F_{ext}(\cdot)$ &  extract features(Figure 1)\\
			drop  & Dropout ratio \\
			\bottomrule
		\end{tabular}%
		\label{tab:addlabel}%
		\caption{The Additional explanation}
		\vspace{-0.2cm}
	\end{table*}%

	\section{Gini importance}
	We present the algorithm of computing the Gini importance used in our paper in Algorithm~\ref{alg:gini}.
	
	\begin{algorithm}[h]  
		\begin{algorithmic}[1]  
			\Require  
			$H$: composed of $h_1$,$h_2$,...,$h_t$ from stage $i$;
			
			{\color[RGB]{162,205,133}{\#The size of $H$ is ($b_z \times c_z \times f_z$)}}
			
			{\color[RGB]{162,205,133}{\#$b_z$ denotes the batch size of $h_t$}}
			
			{\color[RGB]{162,205,133}{\#$c_z$ denotes the number of the feature maps' channel in current stage}}
			
			{\color[RGB]{162,205,133}{\#$f_z$ denotes the number of layers in current stage}}
			
			\Ensure  
			The hotmap $G$ about the features integration for stage $i$;
			\State initial  $G = \emptyset$ ; 
			
			\For{$j=1$ to $f_z-1$}  
			\State $x \gets [h_1,h_2,...,h_{j-1}]$;
			\State $y  \gets [h_j]$;
			\State $x$ $\gets$ $x$.reshape($b_z$,$(f_z - j) \times c_z$);
			\State RF $\gets$ RandomForestRegressor();
			\State RF.fit($x$,$y$);
			\State Gini\_importances $\gets$ RF.feature\_importances\_;

			{\color[RGB]{162,205,133}{\#The length of Gini\_importance is $(f_z - j) \times c_z$}}

			\State res $\gets$ $\emptyset$; 
			\State $s$ $\gets$ 0; 
			\State cnt $\gets$ 0;
			\For{$k=0$ to $(f_z - j)$}
			\State $s$ $\gets$ $s$ + Gini\_importance($k$);
			\State cnt $\gets$ cnt + 1;
			\If {cnt == $c_z$-1}
			
			\State res.add($s$);
			\State $s$ $\gets$ 0;
			\State cnt $\gets$ 0;
			\EndIf  
			
			$G$.add(res/$\max$(res));
			
			\EndFor
			
			\EndFor  
			
		\end{algorithmic}  
		\caption{Calculate features integration by Gini importance from Random Forest}  
		\label{alg:gini}  
	\end{algorithm}

	\section{Number of parameter of LSTM}
	Suppose the input $y_t$ is of size $N$ and the hidden state vector $h_{t-1}$ is also of size $N$.
	
	\textbf{Standard LSTM}
	As shown in Figure (3) (Left), in the standard LSTM, there requires 4 linear transformation to control the information flow with input $y_t$ and $h_{t-1}$ respectively. The output size is set to be $N$. To simplify the calculation, the bias is omitted. Therefore, for the $y_t$, the number parameters of 4 linear transformation is equal to $4\times n\times n$. Similarly, the number parameters of 4 linear transformation with input $h_{t-1}$ is equal to $4\times n\times n$. The total of parameters equals to $8n^2.$
	
	\textbf{DIA-LSTM} As shown in Figure (3) (Right), there is a linear transformation to reduce the dimension at the beginning. The dimension of input $y_t$ will reduce from $N$ to $N/r$ after the first linear transformation. The number of parameters for the linear transformation is equal to $n\times n/r$. Then the output will be passed into 4 linear transformation same as the standard LSTM. the number parameters of 4 linear transformation is equal to $4\times n/r \times n$. Therefore, for input $y_t$ and reduction ratio $r$, the number of parameters is equal to $5n^2/r$. Similarly, the number of parameters with input $h_{t-1}$ is the same as that concerning $y_t$. The total of parameters equals to $10n^2/r.$
	
\end{document}